\documentclass[letterpaper]{article} 
\usepackage{aaai2026}
\usepackage{times}  
\usepackage{helvet}  
\usepackage{courier}  
\usepackage[hyphens]{url}  
\usepackage{graphicx} 
\usepackage{natbib}  
\setcitestyle{authoryear,open={(},close={)}}
\usepackage{caption} 
\usepackage{algorithm}
\usepackage{algorithmic}
\usepackage{amsfonts}
\usepackage{multirow}
\usepackage{graphicx}
\usepackage{color}
\usepackage{subcaption}
\usepackage{booktabs}
\usepackage{amsmath}
\usepackage{newfloat}
\usepackage{listings}
\DeclareCaptionStyle{ruled}{labelfont=normalfont,labelsep=colon,strut=off} 
\floatstyle{ruled}
\newfloat{listing}{tb}{lst}{}
\floatname{listing}{Listing}
\title{AILoRA: Function-Aware Asymmetric Initialization for Low-Rank Adaptation of Large Language Models}

\author {
    Xiaoshuang Ji\textsuperscript{\rm 1, \rm 2, \rm3},
    Zhendong Zhao\textsuperscript{\rm 1, \rm 2, \rm3},
    Xiaoyan Gu\textsuperscript{\rm 1, \rm 2, \rm3},
    Xiaojun Chen\textsuperscript{\rm 1, \rm 2, \rm3},
    Xin Zhao\textsuperscript{\rm 1, \rm 2, \rm3},
    Zeyao Liu\textsuperscript{\rm 1, \rm 2, \rm3},
}
\affiliations {
    \textsuperscript{\rm 1}Institute of Information Engineering,Chinese Academy of Sciences, Beijing, China\\
    \textsuperscript{\rm 2}School of Cyber Security, University of Chinese Academy of Sciences, Beijing, China\\
    \textsuperscript{\rm 2}State Key Laboratory of Cyberspace Security Defense, Beijing, China\\
    jixiaoshuang@iie.ac.cn
}

\usepackage{bibentry}

\begin{document}

\maketitle

\begin{abstract}
Parameter-efficient finetuning (PEFT) aims to mitigate the substantial computational and memory overhead involved in adapting large-scale pretrained models to diverse downstream tasks.
Among numerous PEFT strategies, Low-Rank Adaptation (LoRA) has emerged as one of the most widely adopted approaches due to its robust empirical performance and low implementation complexity.
In practical deployment, LoRA is typically applied to the $W^Q$ and $W^V$ projection matrices of self-attention modules, enabling an effective trade-off between model performance and parameter efficiency.
While LoRA has achieved considerable empirical success, it still encounters challenges such as suboptimal performance and slow convergence.
To address these limitations, we introduce \textbf{AILoRA}, a novel parameter-efficient method that incorporates function-aware asymmetric low-rank priors.
Our empirical analysis reveals that the projection matrices $W^Q$ and $W^V$ in the self-attention mechanism exhibit distinct parameter characteristics, stemming from their functional differences. Specifically, $W^Q$ captures task-specific semantic space knowledge essential for attention distributions computation, making its parameters highly sensitive to downstream task variations. In contrast, $W^V$ encodes token-level feature representations that tend to remain stable across tasks and layers.
Leveraging these insights, AILoRA performs a function-aware initialization by injecting the principal components of $W^Q$ to retain task-adaptive capacity, and the minor components of $W^V$ to preserve generalizable feature representations. This asymmetric initialization strategy enables LoRA modules to better capture the specialized roles of attention parameters, thereby enhancing both finetuning performance and convergence efficiency.
Extensive experiments on multiple large language models and diverse natural language tasks demonstrate the consistent superiority of AILoRA over existing PEFT approaches.
\end{abstract}


\section{Introduction}
Large Language Models (\citealp[LLMs]{brown2020language, ouyang2022training, mao2024dora}), pretrained on large-scale text corpora, have exhibited remarkable generalization capabilities and broad applicability across a wide range of NLP tasks \citep{zheng2023judging}, including mathematical reasoning~\citep{wang2024math} and question answering~\citep{ivison2023camels}.
In practice, full finetuning remains a widely adopted approach for adapting large language models to specific downstream tasks.
However, the substantial computational and memory costs of full finetuning limit its applicability in real-world scenarios.
\begin{figure}[t]
  \hspace{-5pt}
  \begin{subfigure}[t]{0.2\textwidth}
    \centering
    \includegraphics[width=\linewidth]{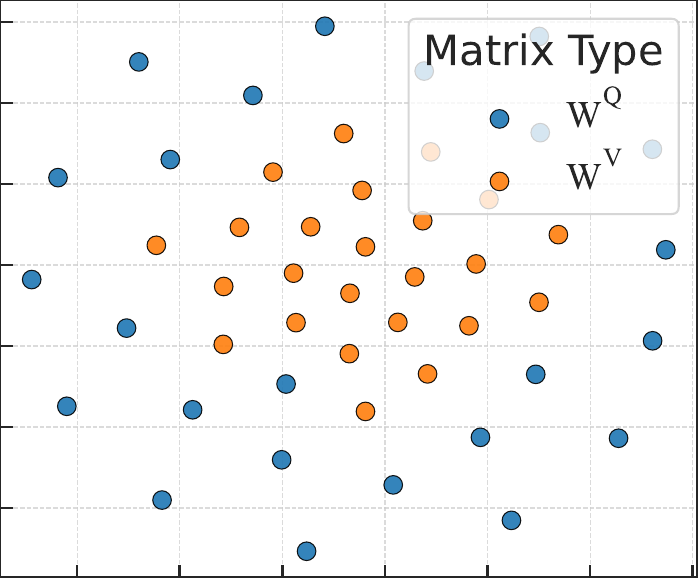}
    \caption{Before finetuning}
  \end{subfigure}
  \begin{subfigure}[t]{0.2\textwidth}
    \centering
    \includegraphics[width=\linewidth]{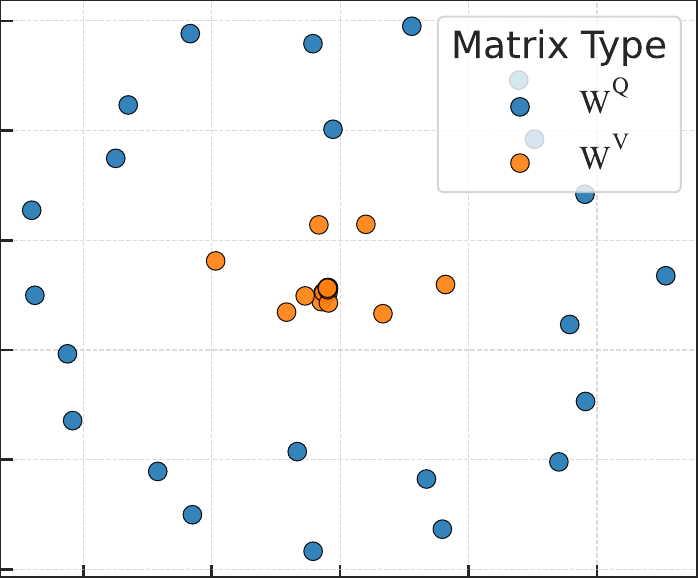}
    \caption{After finetuning}
  \end{subfigure}
  \begin{subfigure}[t]{0.233\textwidth}
    \centering
    \includegraphics[width=\linewidth]{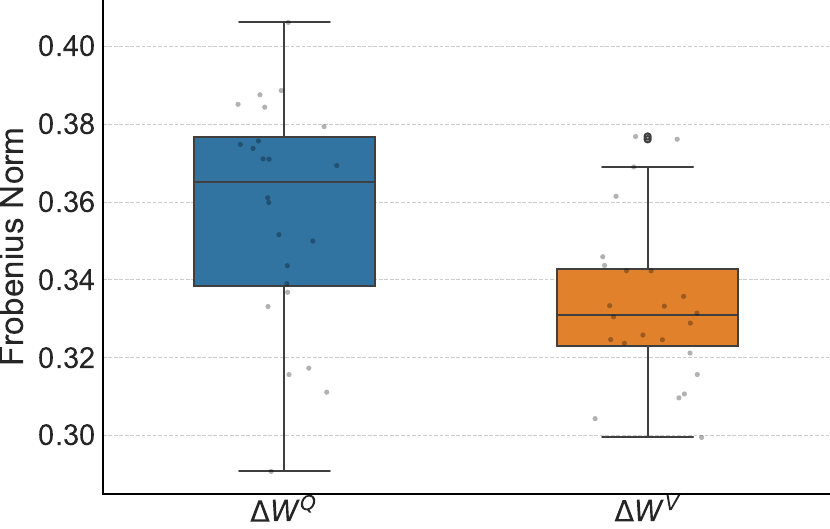}
    \caption{Box plot on CoLA dataset}
  \end{subfigure}
  \begin{subfigure}[t]{0.233\textwidth}
    \centering
    \includegraphics[width=\linewidth]{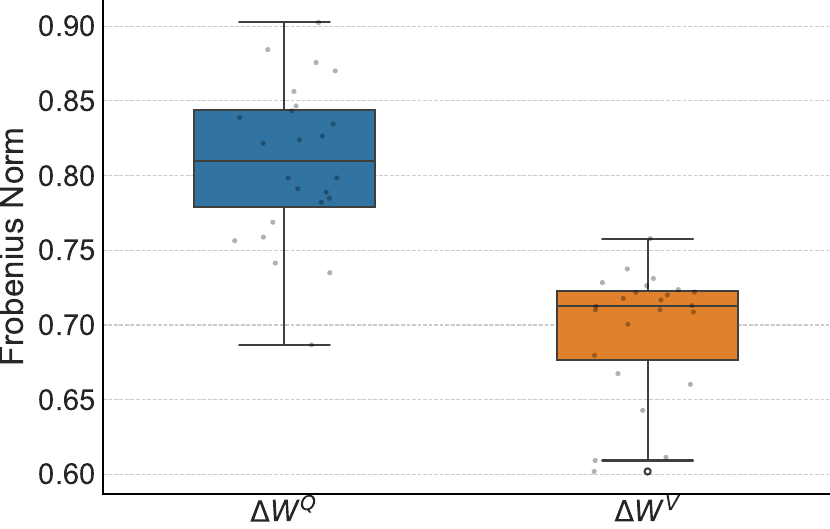}
    \caption{Box plot on SST-2 dataset}
  \end{subfigure}
  \caption{Comparative analysis of the $W^Q$ and $W^V$ projection matrices in the self-attention mechanism.
Figures (a) and (b) visualize the $W^Q$ and $W^V$ matrices across all layers of RoBERTa-large (24 decoder layers) before and after fine-tuning on the CoLA dataset, using t-SNE for dimensionality reduction. Each point represents a projection matrix from a specific layer.
Figures (c) and (d) report the Frobenius norms of the weight updates $\Delta W^Q$ and $\Delta W^V$ after fine-tuning on the CoLA and SST-2 datasets, respectively.}
  \label{fig-inspiration}
\end{figure}
For instance, finetuning a LLaMA-65B model requires over 780GB of GPU memory~\citep{dettmers2024qlora}, and mandates storing a full set of model parameters for each downstream task.
\par
To address these challenges, parameter-efficient finetuning (PEFT; \citealp{hu2023llm}) has emerged as an effective alternative to full finetuning for adapting large-scale pretrained models to downstream tasks, typically by freezing most model parameters and updating only a small number of trainable components.
Recents years have witnessed the rapid emergence of numerous PEFT methods, including Adapter tuning~\citep{houlsby2019parameter}, Prefix-tuning~\citep{li2021prefix}, LoRA~\citep{hu2021lora}, and BitFit~\citep{zaken2022bitfit}.
Among these approaches, Low-Rank Adaptation (LoRA) has received particular attention due to its strong empirical performance and high parameter efficiency.
Specifically, LoRA reduces finetuning overhead by decomposing the weight updates into the product of two randomly initialized low-rank matrices, which are typically applied to the $W^Q$ and $W^V$ matrices of self-attention modules in practice to achieve a balance between parameter efficiency and model performance.
Nonetheless, an increasing number of empirical studies has shown that such random initialization often fails to yield optimal adaptation performance in downstream applications.
With the aim of improving the initialization of LoRA modules, PiSSA~\citep{meng2024pissa} and MiLoRA~\citep{wang2024milora} utilize heuristically selected singular components from pretrained weights to initialize the low-rank matrices, aiming to enhance adaptation performance.
However, they do not take into account the distinct functional roles of the attention projection matrices $W^Q$ and $W^V$, and a uniform singular value-based initialization may still fall short of achieving optimal performance on downstream tasks. This limitation becomes more pronounced as model size increases, where the oversimplified initialization strategy struggles to accommodate the growing complexity and functional heterogeneity of large-scale models.
\par
To overcome the aforementioned limitations and improve the initialization scheme in low-rank adaptation, we first examine the functional differences and parameter behaviors of the attention projection matrices $W^Q$ and $W^V$ in self-attention.
Prior work~\citep{vaswani2017attention, clark2019does} has shown that the $W^Q$ projection matrices in self-attention generate query vectors that guide attention over the semantic space, playing a key role in semantic alignment. In contrast, $W^V$ produces value vectors that encode token-level features and are aggregated via attention to produce the final output representations.
Inspired by the aforementioned perspective, we conduct a comparative analysis of the $W^Q$ and $W^V$ projection matrices to investigate their parameter distribution patterns and variation trends in relation to downstream tasks.
As illustrated in Figure~\ref{fig-inspiration}, two notable phenomena can be observed: (1) the distribution of the $W^Q$ matrices exhibits a high degree of dispersion after finetuning, and the Frobenius norms of $\Delta W^Q$ are relatively large, suggesting that $W^Q$ across different layers captures diverse semantic information and is highly sensitive to downstream tasks; (2) in contrast, the $W^V$ matrices display a highly concentrated and layer-consistent distribution, and the relatively small Frobenius norms of $\Delta W^V$ indicate more stable and task-invariant encoding behavior, reflecting their role in representing generalizable token-level features.
\par
Inspired by the above analysis, we propose Function-aware \textbf{A}symmetric \textbf{I}nitialization for \textbf{Lo}w-\textbf{R}ank \textbf{A}daption (AILoRA), a parameter-efficient finetuning method that introduces an asymmetric initialization strategy to better align with the distinct functional roles of projection matrices in self-attention mechanism.
Specifically, we perform singular value decomposition (SVD) on the pretrained $W^Q$ and $W^V$ matrices, and utilize the dominant singular components (those with the largest singular values) of $W^Q$ and the minor components (associated with the smallest singular values) of $W^V$ to initialize their respective LoRA modules.
This asymmetric initialization offers two key advantages: (1) it enables the LoRA modules of $W^Q$ to rapidly adapt to the semantic space of downstream tasks and extract task-relevant semantic features, thereby facilitating more domain-sensitive attention computation; (2) it allows the LoRA modules of $W^V$ to refine task-specific representations while preserving the generalizable feature encoding capabilities acquired during pretraining.
We conduct comprehensive experiments across various model architectures, parameter scales, and datasets from diverse downstream tasks. The results demonstrate that AILoRA consistently outperforms mainstream PEFT methods in both performance and convergence speed.
The main contributions are summarized as follows:
\begin{itemize}
    \item To enhance the effectiveness of LoRA, we are the first to leverage the functional asymmetry of the self-attention projection matrices: $W^Q$ captures task-sensitive semantic information essential for attention distribution, whereas $W^V$ encodes more stable token-level features. 
    \item Based on our empirical observations, we propose AILoRA, a novel PEFT method that introduces a function-aware asymmetric initialization strategy for LoRA modules, effectively striking a better balance between task-specific adaptability and the retention of pretrained knowledge.
    \item Comprehensive experiments across diverse model architectures, parameter scales, and downstream tasks demonstrate that AILoRA consistently surpasses existing PEFT baselines, while significantly accelerating convergence.
\end{itemize}
\section{Related Works}
\subsection{Parameter-efficient Finetuning}
\begin{figure*}[ht]
  \includegraphics[width=\linewidth]{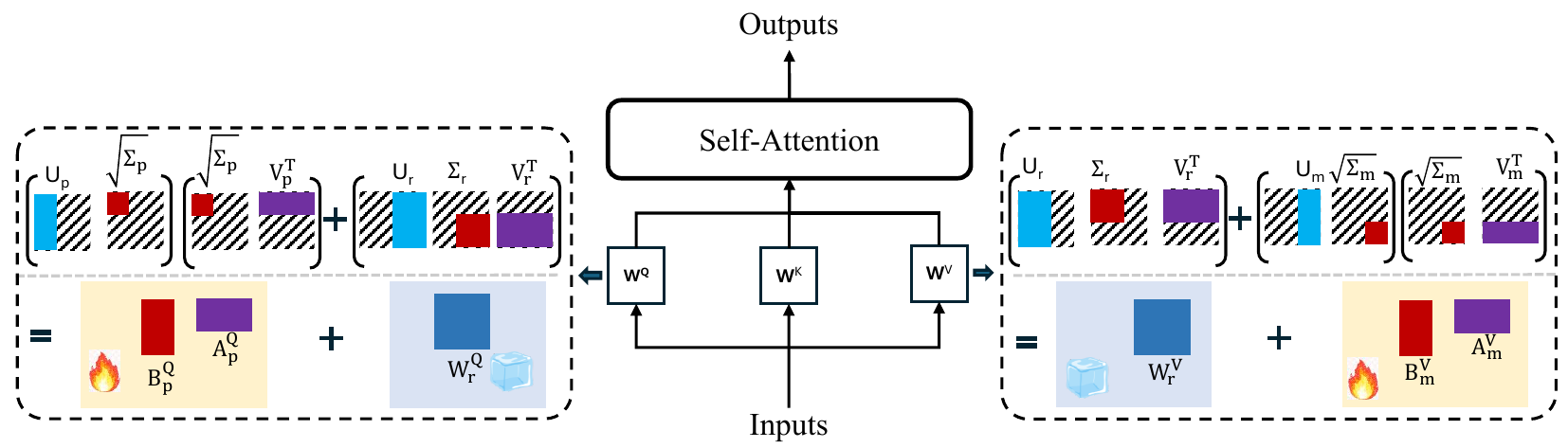}
  \caption {AILoRA first performs SVD on the $W^Q$ and $W^V$ matrices. For the $W^Q$ matrices, the principal components are used to initialize the LoRA modules while keeping the remaining components frozen.In contrast, the LoRA modules of $W^V$ are initialized using the minor components, with the remaining components fixed.}
  \label{fig-framework}
\end{figure*}
Despite its success across numerous tasks, finetuning still has several limitations.
Notably, finetuning requires updating all parameters of pretrained models, which is impractical given the explosive growth of parameter amounts.
Recent years have witnessed the rise of parameter-efficient finetuning methods, known as PEFT.
PEFT techniques freeze most parameters and update only a small set of parameters to reduce computing resource consumption without compromising model performance.
There are three mainstream classes of PEFT methods: addition-based, selection-based, and reparametrization-based \cite{lialin2023scaling}.
The addition-based PEFT methods freeze pretrained weights and inject trainable parameters or modules, such as Adapter tuning \cite{houlsby2019parameter}, Prefix-tuning \cite{li2021prefix} and Prompt tuning \cite{lester2021power}.
The selection-based PEFT methods select a subset of parameters and freeze the rest, including BitFit \cite{zaken2022bitfit} and FAR \cite{vucetic2022efficient}.
The reparametrization-based methods introduce reparametrization to reduce trainable parameters, such as LoRA and KronA \cite{edalati2022krona}.
Building upon these methods, numerous variants have subsequently emerged, including P-tuning \citep{liu2023gpt}, DoRA \citep{liu2024dora}, QLoRA \citep{dettmers2024qlora} and LoftQ \citep{li2024loftq}.
\subsection{Low-Rank Adaptation}
The low-rank adaptation (LoRA) is one of the most widely adopted PEFT techniques, grounded in the core assumption that the weight updates necessary for downstream task adaptation are intrinsically low-rank.
Consequently, LoRA employs the product of two low-rank matrices, $A \in \mathbb{R}^{r \times n}$ and $B \in \mathbb{R}^{m \times r}$ to approximate the weight updates of $W \in \mathbb{R}^{m \times n}$.
The model parameters can be expressed as 
\begin{equation}
    W = W_0+\Delta W =W_0 + \frac{\alpha}{r}BA,
    \label{eq:lora}
\end{equation}
where $W_0$ and $\Delta W$ denote the pretrained weights and weight updates, respectively.
The scaling factor $\alpha$ is used to facilitate the optimization process, and $r$ is the rank of two low-rank matrices ($r \ll min(m, n) $).
The $B$ matrix is initialized to all zero, while the $A$ matrix adopts a random Gaussian distribution initialization.
This initialization strategy ensures that $\Delta W =0$ at the beginning, implying no deviation from the pretrained weights.
During finetuning, the pretrained weight $W_0$ keeps frozen and only the two low-rank matrices $A$ and $B$ are trainable.
\section{Methodology}
In this section, we present the details of the proposed method.
Motivated by the observation that different projection matrices in the self-attention mechanism fulfill distinct semantic roles, we hypothesize that adopting a matrix-specific initialization strategy can better exploit their respective capacities.
As highlighted in LASER \citep{sharma2023truth}, the minor singular components of weight matrices contain noisy or long-tail information, while the principal singular components capture essential features across tasks.
Consequently, we propose the function-aware \textbf{A}symmetric \textbf{I}nitialization for \textbf{Lo}w-\textbf{R}ank \textbf{A}daptation based on the unique properties of different singular components.
The framework of AILoRA is illustrated in Figure~\ref{fig-framework}.
At first, AILoRA applies the SVD technique to the pretrained weight matrices $W^Q$ and $W^V$ $\in \mathbb{R}^{m \times n}$.
The SVD result is $W=U\Sigma V^T$, where $U = [u_1,u_2,\dots,u_m] \in \mathbb{R}^{m \times m}$ and $V =[v_1,v_2,\dots, v_n]\in \mathbb{R}^{n \times n}$ are the singular matrices with orthonormal columns and $V^T$ is the transpose of $V$.
$\Sigma \in \mathbb{R}^{m \times n}$ is a diagonal matrix, where diagonal elements are the singular values arranged in descending order.
\begin{table*}[t]
\centering
\caption{\label{glue}
    Results on the GLUE benchmark. The best results are shown in \textbf{bold}.
}
\begin{tabular}{cccccccccccc}
\toprule
\textbf{Model} & \textbf{Method} & \textbf{CoLA} & \textbf{MNLI} & \textbf{MRPC} & \textbf{QNLI} & \textbf{QQP}  & \textbf{RTE}  & \textbf{SST}-2 & \textbf{STS}-B & \textbf{Avg} \\ \midrule
\multirow{4}{*}{\begin{tabular}[c]{@{}c@{}}RoBERTa-large\\ (335M)\end{tabular}}   
& LoRA & 67.9 & 90.2 & 92.7 & 94.3 & \textbf{88.9} & 85.8 & 96.2 & \textbf{91.8} & 88.5 \\ & PiSSA & 67.8 & \textbf{90.3} & 92.1 & \textbf{94.8} & \textbf{88.9} & 85.0 & 96.1 & 91.7 & 88.3 \\& MiLoRA & 68.4 & 90.2 & 93.0 & \textbf{94.8} & 88.8 & 85.4 & 96.2 & 91.7 & 88.6 \\
& AILoRA & \textbf{69.3} & \textbf{90.3} & \textbf{93.5} & 94.7 & 88.8 & \textbf{86.4} & \textbf{96.3} & \textbf{91.8} & \textbf{88.9}\\ \midrule
\multirow{4}{*}{\begin{tabular}[c]{@{}c@{}}DeBERTa-v3-base\\ (184M)\end{tabular}} 
& LoRA   & 69.5 & 89.8 & 92.8 & \textbf{94.2} & \textbf{89.0} & 85.6 & 96.0 & 90.9 & 88.5 \\
& PiSSA  & 68.8 & 89.0 & 92.2 & 94.1 & 88.5 & 84.3 & 96.0 & 90.9 & 88.0 \\
& MiLoRA & 68.8 & 89.5 & 92.8 & \textbf{94.2} & 88.9 & \textbf{85.8} & 95.8 & \textbf{91.2} & 88.4\\
& AILoRA   & \textbf{69.7} & \textbf{90.0} & \textbf{92.9} & \textbf{94.2} & 88.9 & 85.4 & \textbf{96.1} & \textbf{91.2} & \textbf{88.6}\\ \bottomrule
\end{tabular}
\end{table*}
Then, AILoRA uses the SVD results to initialize the LoRA modules.
\par
For the $W^Q$ matrices, AILoRA utilizes the largest $r$ singular values and their corresponding singular vectors to form the principal low-rank matrices $A_p^Q$ and $B_p^Q$, which can be formulated as:
\begin{equation}
\label{q}
\left\{
\begin{aligned}
B_p^Q = U_{[:,:r]}\Sigma_{[:r,:r]}^{1/2} \in \mathbb{R}^{m \times r} & , \\
A_p^Q = \Sigma_{[:r,:r]}^{1/2}V^T_{[:,:r]} \in \mathbb{R}^{r \times n} & .
\end{aligned}
\right.
\end{equation}
And the remaining components are used to construct the residual matrices $W_r^Q$, which are frozen during finetuning:
\begin{equation}
    W_r^Q = U_{[:,r:]}\Sigma_{[r:,r:]}V^T_{[:,r:]} \in \mathbb{R}^{m \times n}.
    \label{eq:q}
\end{equation}
The matrix slicing notations used above are consistent with those in Python, in which [:r] denotes the first r dimensions.
The low-rank matrices $A_p^Q$ and $B_p^Q$ can be multiplied to obtain the full-size principal matrices $W_p^Q$ and the $W^Q$ matrices can be formed:
\begin{equation}
    \begin{split}
        W^Q&=W_p^Q+W_r^Q=B_p^QA_p^Q+W_r^Q. \\
    \end{split}
\end{equation}
The LoRA modules of the $W^Q$ matrices contain knowledge that significantly influences attention computation.
By training the principal components, the $W^Q$ matrices can swiftly adapt to the semantic space of downstream tasks and perform a more domain-oriented computation of attention distribution, while less critical knowledge is preserved in the $W_r^Q$ matrices to reduce computational overhead.
\par
For the $W^V$ matrices, AILoRA utilizes the smallest $r$ singular values and their corresponding singular vectors to construct the minor low-rank matrices $A_m^V$ and $B_m^V$.
The remaining components construct the residual matrices $W_r^V$, which is kept frozen during finetuning:
\begin{equation}
\label{v}
\left\{
\begin{array}{c}
B_m^V = U_{[:,-r:]}\Sigma_{[-r:,-r:]}^{1/2} \in \mathbb{R}^{m \times r}, \\
A_m^V = \Sigma_{[-r:,-r:]}^{1/2}V^T_{[:,-r:]} \in \mathbb{R}^{r \times n}, \\
W_r^V = U_{[:,:-r]}\Sigma_{[:-r,:-r]}V^T_{[:,:-r]} \in \mathbb{R}^{m \times n}.
\end{array}
\right.
\end{equation}
The matrix slicing notations [-r:] denote the last r dimensions.
Similarly, the low-rank matrices $A_m^V$ and $B_m^V$ are used to reconstruct  the full-size minor matrices $W_m^V$ and the $W^V$ matrices can be formed:
\begin{equation}
    \begin{split}
        W^V&=W_m^V+W_r^V=B_m^VA_m^V+W_r^V. \\
    \end{split}
\end{equation}
The minor components of the $W^V$ matrices, encapsulating less critical knowledge, are assigned to the low-rank matrices $A_m$ and $B_m$.
The optimization process enables the LoRA modules to master feature representations tailored to downstream tasks and mitigate the impact of noise.
And the remaining components in the $W_r^V$ matrices remain unchanged to preserve knowledge acquired by pretraining.
\par
At last, the LoRA modules get updated and the residual matrices are kept frozen during finetuning.
The design of AILoRA similarly ensures no deviation from the pretrained weights at the beginning of training.
\section{Experiments}
To assess the effectiveness of AILoRA, we conduct extensive experiments on both Natural Language Understanding (NLU) and Natural Language Generation (NLG) tasks.
The baselines include LoRA, PiSSA and MiLoRA.
All experiments are performed on a single NVIDIA A100 GPU unless otherwise specified.
\begin{itemize}
    \item LoRA utilizes the products of two low-rank matrices $A$ and $B$ to approximate the weight updates $\Delta W$.
    At the beginning of finetuning, the matrix $A$ is initialized with random Gaussian values and B is initialized to all zero, ensuring that $\Delta W=0$. 
    \item Considering that the principal components capture the essential knowledge of a matrix, PiSSA solely initializes the low-rank matrices $A$ and $B$ with the principal singular values and vectors of the pretrained weight $W$.
    \item In contrast, MiLoRA merely utilizes the less-optimized minor singular values and vectors of the pretrained weight $W$ to initialize the low-rank matrices $A$ and $B$ based on the view that the minor singular components of weight matrices may contain noisy or long-tail knowledge.
\end{itemize}
\subsection{Experiments on NLU Tasks}
\textbf{Models and Datasets} We finetune RoBERTa-large \citep{liu2019roberta}, an encoder-only model consisting of 24 layers, on the GLUE benchmark \citep{wang2018glue} and SQuAD datasets \citep{rajpurkar2016squad}.
The GLUE benchmark comprises nine natural language understanding tasks, covering single-sentence classification, similarity and paraphrase, and inference tasks.
Consistent with prior researches, we exclude the WNLI task.
Evaluation metrics also follow prior works: CoLA is evaluated using Matthew’s Correlation, STS-B with Spearman’s correlation coefficient, MRPC and QQP with F1 score, and the remaining tasks are evaluated using accuracy.
The SQuAD datasets include two versions, SQuAD v1.1 and SQuAD v2.0, for which we report the Exact Match (EM) ratio and F1 score.
Additionally, the encoder-only DeBERTa-v3-base with 12 layers \citep{he2021debertav3} is also used on the GLUE benchmark for further comparison.
\begin{table}
\centering
\caption{\label{squad}
    Results with RoBERTa-large on SQuAD datasets. The best results are shown in \textbf{bold}
}
\begin{tabular}{cccc}
\toprule
\textbf{Dataset}                     & \textbf{Method} & \textbf{EM} & \textbf{F1} \\ \midrule
\multirow{4}{*}{SQuAD v1.1} & LoRA   & \textbf{88.6} & \textbf{94.4} \\
                            & PiSSA  & 88.4 & 94.3 \\
                            & MiLoRA & 88.5 & \textbf{94.4} \\
                            & AILoRA   & 88.4 & 94.3 \\ \midrule
\multirow{4}{*}{SQuAD v2.0} & LoRA   & 78.0 & 81.2 \\
                            & PiSSA  & 77.4 & 81.1 \\
                            & MiLoRA & 77.5 & 81.2 \\
                            & AILoRA   & \textbf{78.5} & \textbf{82.2} \\ \bottomrule
                            
\end{tabular}
\end{table}
\\
\textbf{Implementations Details} 
We use the implementation of transformers\footnote{https://github.com/huggingface/transformers} for all NLU tasks.
For the GLUE benchmark, the rank of low-rank matrices is uniformly set to 8 across all methods and datasets
We conduct a grid search over learning rates in \{3e-5, 4e-5, 5e-5, 2e-4, 3e-4, 4e-4\} and report the best results.
Batch size, epoch number and other hyperparameters are consistent with PiSSA.
For the SQuAD datasets, the numbers of epochs for SQuAD v1.1 and v2.0 are 3 and 2, respectively, with learning rates searched over \{1e-4, 2e-4, 3e-4, 4e-4\}.
The rank of low-rank matrices is similarly fixed to 8.
All experiments are repeated 5 times using random seeds, and the reported results are averaged over these runs.
\\
\textbf{Results} We present the results on the GLUE benchmark and SQuAD datasets in Table~\ref{glue} and Table~\ref{squad}, respectively.
As shown in Table~\ref{glue}, our proposed AILoRA achieves the best performance on 12 out of 16 tasks and also yields the highest average score across all tasks.
AILoRA exceed the best baseline by 0.8 points on the challenging textual entailment task RTE.
In Table~\ref{squad}, AILoRA demonstrates competitive performance compared to baseline methods on SQuAD v1.1 with negligible performance gaps.
On more challenging SQuAD v2.0, AILoRA outperforms all baselines, yielding improvements of 0.5 and 1.0 points in EM and F1 scores, respectively.
In conclusion, these results highlight that AILoRA demonstrates enhanced adaptability to downstream tasks, which contributes to its superior performance across a variety of NLU tasks.
\subsection{Experiments on NLG Tasks}
\begin{table}
\centering
\caption{\label{sum}
    Results with BART-large on summarization tasks. We report R-1/2/L scores. The best results are shown in \textbf{bold}.
}
\begin{tabular}{ccccccc}
\toprule
\textbf{Method} & \multicolumn{3}{c}{\textbf{XSUM}}           & \multicolumn{3}{c}{\textbf{CNN/DailyMail}}  \\ \midrule
LoRA   & \multicolumn{3}{c}{40.46 / 17.55 / 32.36} & \multicolumn{3}{c}{42.73 / 19.75 / 29.17} \\
PiSSA  & \multicolumn{3}{c}{40.63 / \textbf{17.63} / \textbf{32.51}}               & \multicolumn{3}{c}{42.74 / 19.65 / 29.18} \\
MiLoRA & \multicolumn{3}{c}{40.31 / 17.35 / 32.15} & \multicolumn{3}{c}{42.84 / 19.78 / 29.17}               \\
AILoRA & \multicolumn{3}{c}{\textbf{40.66} / 17.61 / \textbf{32.51}}               & \multicolumn{3}{c}{\textbf{42.91} / \textbf{19.79} / \textbf{29.24}}               \\ \bottomrule
\end{tabular}
\end{table}
\begin{table}
\centering
\caption{\label{math}
    Results with LLaMA2-7B on math reasoning tasks. The best results are shown in \textbf{bold}.
}
\begin{tabular}{cccc}
\toprule
\textbf{Method} & \textbf{GSM8K} & \textbf{MATH} & \textbf{Avg.} \\ \midrule
LoRA   & 49.7 & 7.0 & 28.4 \\
PiSSA  & \textbf{53.4} & 8.4 & 30.9 \\
MiLoRA & 49.4 & 7.3 & 28.4 \\
AILoRA   & \textbf{53.4} & \textbf{8.9} & \textbf{31.1} \\ \bottomrule
\end{tabular}
\end{table}
\textbf{Models and Datasets} We finetune BART-large \citep{lewis2020bart}, which adopts an encoder\&decoder architecture with 12 encoder layers and 12 decoder layers, on two summarization datasets: XSum \citep{narayan2018don} and CNN/DailyMail \citep{hermann2015teaching}, and evaluate model performance using Rouge 1/2/L scores (R-1/2/L, \citep{lin2004rouge}).
In addition, we finetune the decoder-only LLaMA2-7B with 32 layers \citep{touvron2023llama} on math reasoning using the training set of MetaMathQA-385K \citep{yu2023metamath}.
\begin{figure}[t]
  \begin{subfigure}[t]{0.46\textwidth}
    \centering
    \includegraphics[width=\linewidth]{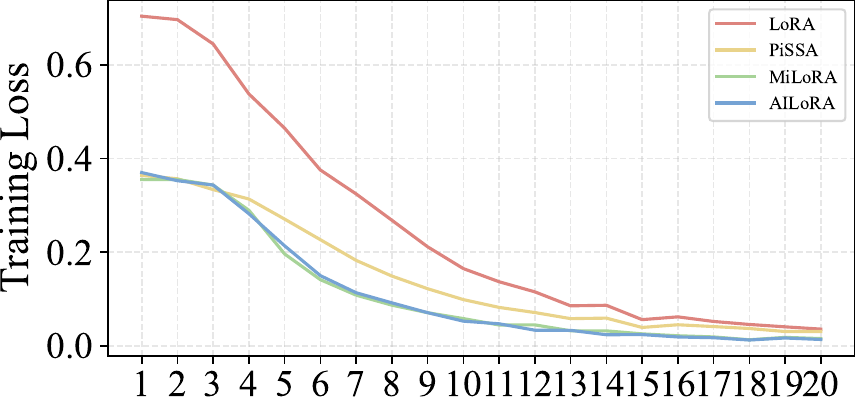}
    \caption{Training loss over epochs}
    \label{loss_convergence}
  \end{subfigure}
  \begin{subfigure}[t]{0.46\textwidth}
    \centering
    \includegraphics[width=\linewidth]{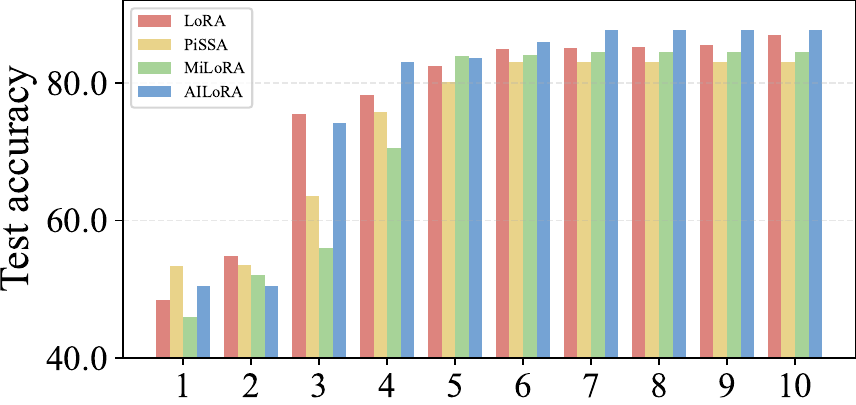}
    \caption{Test accuracy over epochs}
    \label{accuracy_convergence}
  \end{subfigure}
  \caption {The training loss and accuracy over the epochs of AILoRA and baselines.}
  \label{convergence}
\end{figure}
Evaluation is conducted on the test sets of GSM8K \citep{cobbe2021training} and MATH \citep{hendrycks2021measuring}, where we report the Exact Match (EM) ratio.
\\
\textbf{Implementation Details} For summarization tasks, we use the implementation of transformers and follow the setting of AdaLoRA \citep{zhang2023adaptive}, setting the rank of low-rank matrices to 8 and training epochs to 15.
\begin{figure*}[t]
  \begin{subfigure}[t]{0.68\textwidth}
    \centering
    \includegraphics[width=\linewidth]{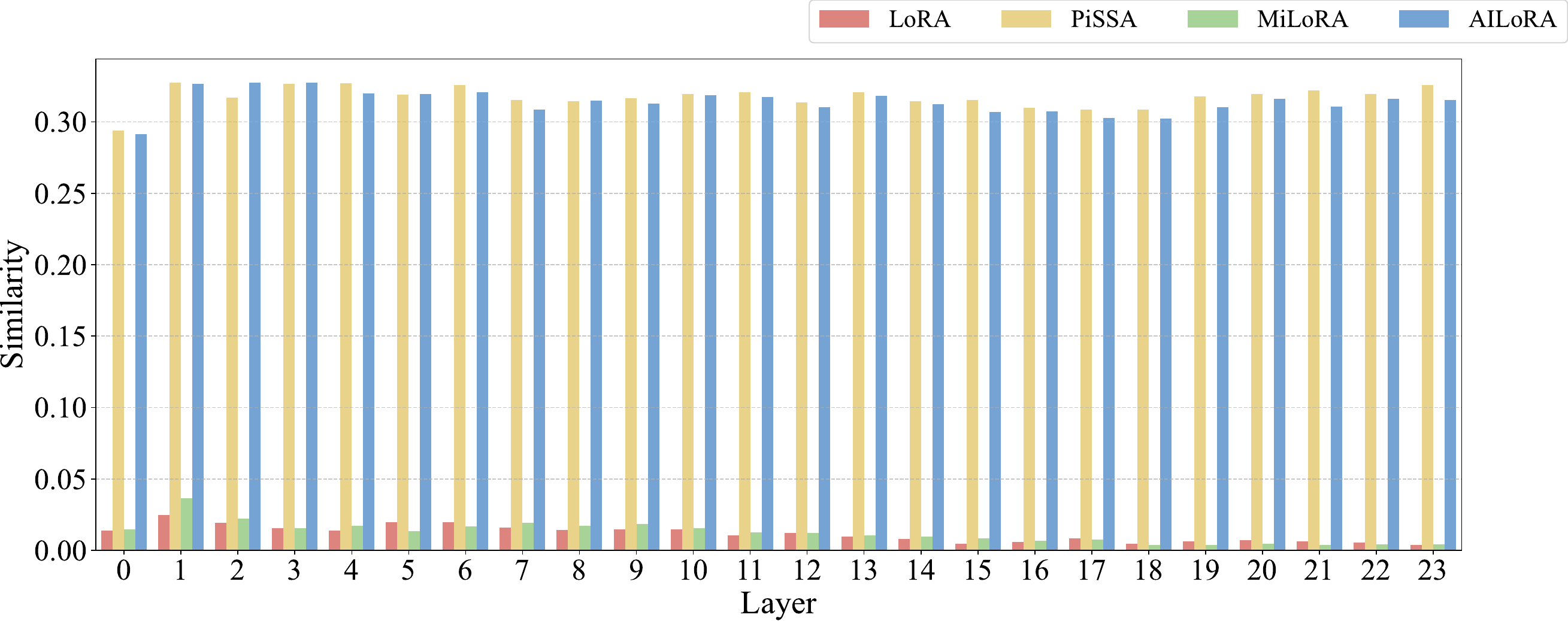}
    \caption{Similarity Between low-rank matrices of $W^Q$ and $\Delta W^Q$ across layers}
    \label{similarity_q}
  \end{subfigure}
  \hspace{-0.042\textwidth}
  \begin{subfigure}[t]{0.35\textwidth}
    \centering
        \raisebox{6pt}{\includegraphics[width=\linewidth]{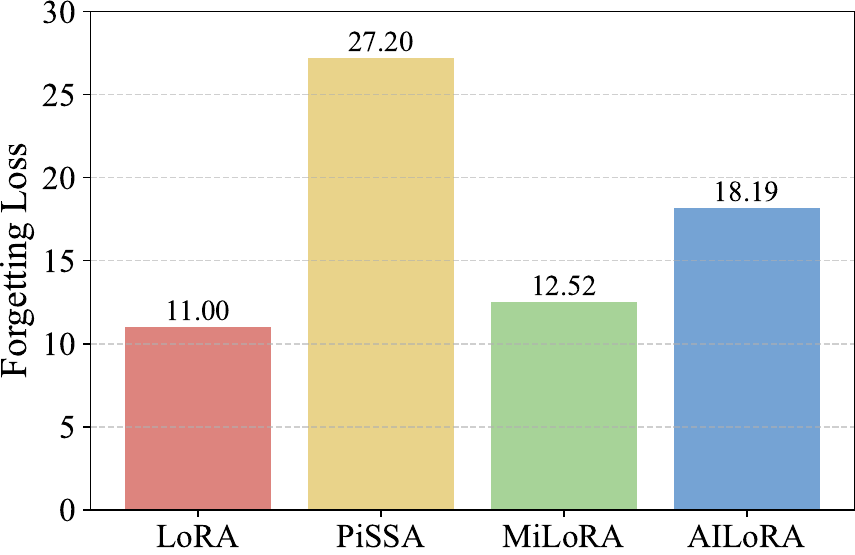}}
    \caption{Forgetting loss of AILoRA and baselines}
    \label{forget}
  \end{subfigure}
  \caption{Experiments on function-aware enhancement of $W^Q$ and $W^V$.}
  \label{fig-similarity}
\end{figure*}
For XSum, we set the beam length as 8 and batch size as 64, while for CNN/DailyMail, we set the beam length as 4 and batch size 32.
We conduct a grid search over learning rates in \{5e-5, 1e-4, 5e-4\} and report the best results.
For math reasoning, we use the implementation of MetaMath \citep{yu2023metamath}\footnote{https://github.com/meta-math/MetaMath}.
The AdamW optimizer \citep{loshchilov2017decoupled} is employed with a learning rate 2e-5, warming up for 3\% steps.
We finetune LLaMA2-7B for 3 epochs and set the rank of low-rank matrices to 64 to accommodate the larger training corpus.
All experiments are repeated 5 times on a single NVIDIA A800 GPU using random seeds to report the average results.
\\
\textbf{Results} Table~\ref{sum} and Table~\ref{math} present the results of summarization and math reasoning, respectively.
On summarization tasks, AILoRA consistently achieves best scores across all matrics, closely matching or surpassing the best-performing baselines.
For math reasoning, AILoRA consistently outperforms all baselines on both GSM8K and MATH dataset, and achieves the best overall average performance.
On the more challenging MATH dataset, AILoRA achieves an EM score of 8.9, exceeding LoRA's 7.0 by 1.9 point, which highlights that AILoRA’s function-aware initialization strategy significantly enhances the model’s downstream tasks adaption.
\subsection{Convergence analysis}
To assess the convergence behavior of AILoRA and baselines, we finetune RoBERTa-large on RTE dataset for 20 epochs with the rank $r$ set to 8.
Both training loss and test accuracy at each epoch are visualized as shown in Figure~\ref{convergence}.
As shown in Figure~\ref{loss_convergence}, we observe that AILoRA consistently maintains the lowest training loss across epochs, indicating more efficient and stable optimization.
Notably, LoRA exhibits a significantly higher loss after the first epoch, which highlights that the standard initialization strategy used in LoRA prevents efficient convergence at early stages.
As illustrated in Figure~\ref{accuracy_convergence}, AILoRA reaches 80\% test accuracy within the first four epochs, faster than all baselines, and ultimately achieves the highest accuracy of 87.7\%.
These results further validate the effectiveness of AILoRA in accelerating convergence and enhancing overall performance.
\subsection{Comparison with More PEFTs}
\begin{table}[t]
\centering
\caption{\label{ablation}
    Comparison with more PEFT methods. The number of trainable parameters is reported to two decimal places. The best results are shown in \textbf{bold}.
}
\setlength{\tabcolsep}{5pt}
\begin{tabular}{cccccc}
\toprule
\textbf{Method} & \textbf{Params} & \textbf{CoLA} & \textbf{MRPC} & \textbf{RTE} & \textbf{STS-B} \\ \midrule
Full FT        & 355.36M & 68.5 & 93.1 & 85.8 & \textbf{92.1} \\ \midrule
Adapter        & 7.40M & 68.5 & 93.1 & \textbf{87.0} & \textbf{92.1} \\
BitFit         & 1.32M & 68.4 & 92.7 & 86.4 & 91.7 \\
DoRA           & 0.84M & 67.3 & 93.2 & 83.6 & 91.8 \\
rsLoRA         & \textbf{0.79M} & 67.1 & 93.0 & 86.8 & 91.8 \\
VeRA           & 1.11M & 68.6 & 93.3 & 83.6 & 91.1 \\ 
AILoRA         & \textbf{0.79M} & \textbf{69.3} & \textbf{93.5} & 86.4 & 91.8  \\ \bottomrule
\end{tabular}
\end{table}
To further assess the effectiveness of AILoRA, we compare AILoRA with full finetuning, classic and novel PEFT methods, including Adapter tuning, BitFit, DoRA, rsLoRA \citep{kalajdzievski2023rank} and VeRA \citep{kopiczko2023vera}.
Specifically, we finetune RoBERTa-large on four tasks of GLUE benchmark: CoLA, MRPC, RTE and STS-B.
As for full finetuning, we adapt the hyperparameter configuration given in the original paper.
Adapter tuning inserts and updates adapter modules into the self-attention and feedforward layers and the bottleneck size of Adapter tuning is set to 64 by default.
BitFit only updates the bias terms of the pretrained model.
DoRA decomposes pretrained weights into two components, direction and magnitude, and only applies LoRA to the direction component to enhance training stability.
rsLoRA divides the LoRA modules by the square root of the rank, facilitating a straightforward finetuning compute/performance trade-off.
VeRA shares a pair of frozen random matrices across all layers and conducts layer-wise adaptation using “scaling vectors”.
The rank of low-rank matrices of VeRA is set to 256, following the configuration used in the original paper.
We repeat all experiments 5 times and report the best average results.
As shown in Table~\ref{ablation}, AILoRA achieves the highest scores on the CoLA and MRPC tasks by updating the fewest parameters, surpassing the best baselines by 0.7 point on CoLA.
On RTE and STS-B tasks, AILoRA still achieves competitive results.
Although Adapter tuning gets the highest scores, it updates $\times$6.7-9.3 more parameters.
Overall, AILoRA achieves an optimal balance between parameter efficiency and model performance.
\subsection{Experiments on Function-Aware Enhancement}
\begin{figure*}[t]
  \begin{subfigure}[t]{0.49\textwidth}
    \centering
        \raisebox{1pt}{\includegraphics[width=\linewidth]{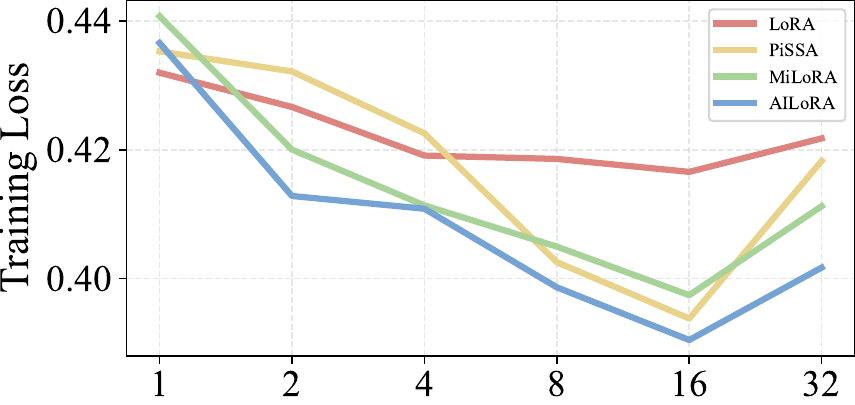}}
    \caption{Training loss under various ranks}
    \label{loss_rank}
  \end{subfigure}
  \hfill
  \begin{subfigure}[t]{0.5\textwidth}
    \centering
    \includegraphics[width=\linewidth]{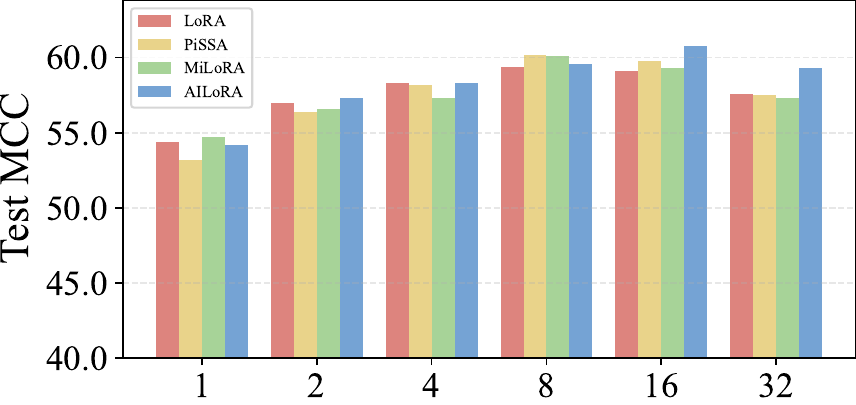}
    \caption{Test MCC under variouse ranks}
    \label{accuracy_rank}
  \end{subfigure}
  \caption {Comparison between AILoRA and baselines across various ranks.}
  \label{fig-rank}
\end{figure*}
To gain deeper insights into AILoRA, we conduct further experiments to investigate the function-aware enhancement of $W^Q$ and $W^V$ matrices.
For $W^Q$ matrices, we assess the similarity between the LoRA modules and $\Delta W$, defined as the difference between the fully-finetuned and pretrained weights.
The analysis follows the method outlined in LoRA.
Specifically, we employ SVD to extract the first r columns of the left singular-vector matrices from both full-size low-rank matrices and $\Delta W$.
Then we compute the subspace similarity using the following metric: $\phi(A, B)=\frac{{\Vert A^TB \Vert}^2_F}{r}$, where ${\Vert \cdot \Vert}_F$ denotes the Frobenius norm.
The value of $\phi(A, B)$ ranges from 0 to 1, with larger values indicating higher subspace similarity.
As presented in Figure~\ref{similarity_q}, the LoRA modules of PiSSA and AILoRA exhibit strong similarity with $\Delta W$, indicating that the knowledge they acquire closely resembles that obtained through full finetuning.
This effect reveals that the LoRA modules of $W^Q$ in PiSSA and AILoRA demonstrates strong adaptability to downstream task-specific semantic space.
For $W^V$ matrices, the cross-entropy loss, the usual next token prediction loss used when training LLMs, is used as the metric for measuring forgetting \citep{kalajdzievski2024scaling}.
The difference is that we assess the divergence between the predicted distributions of the pretrained model and the finetuned model.
To evaluate the forgetting metric, we finetune RoBERTa-large on the CoLA dataset and assess forgetting on the SST-2 test set.
As shown in Figure\ref{forget}, the forgetting losses of AILoRA and MiLoRA are lower than PiSSA, showing that they forget less and better preserve the general linguistic capabilities acquired by pretraining, which benefits from optimizing the minor singular components.
In light of previous experimental results, AILoRA demonstrates superior adaptability to downstream tasks, leading to improved model performance, and exhibits less forgetting of pretrained knowledge.
\subsection{Experiments on Various Ranks}
In this section, we investigate the impact of increasing the rank from 1 to 32, aiming to assess whether AILoRA consistently outperforms baselines across different rank settings.
The experiments are conducted on the CoLA dataset for 2 epochs, and the training loss of the training set and the accuracy on the test set are depicted in Figure~\ref{fig-rank}.
In Figure~\ref{loss_rank}, the training losses of AILoRA are almost the lowest compared to all baselines, indicating the best adaptability to downstream tasks.
Figure~\ref{accuracy_rank} further demonstrates that AILoRA consistently surpasses all baselines under the same parameter budget, highlighting its broad adaptability and scalability.
Notably, when the rank is increased to 32, both the training loss and test Matthews Correlation Coefficient (MCC) exhibit anomalous behavior: the training loss rises and the MCC declines.
This observation highlights that simply increasing the number of trainable parameters does not guarantee improved performance and may even lead to degradation.
\subsection{Experiments on Weight Matrix Selection}
\begin{table}
\centering
\caption{\label{rank}
    Results on different rank settings. The best results are shown in \textbf{bold}.
}
\begin{tabular}{ccccc}
\toprule
\textbf{Rank Settings} & \textbf{CoLA} & \textbf{MRPC} & \textbf{RTE} & \textbf{STS-B} \\ \midrule
$r_q$=$r_v$=8   (AILoRA) & \textbf{69.3} & \textbf{93.5} & \textbf{86.4} & 91.8 \\
$r_q$=16                 & 62.1 & 92.2 & 82.0 & 90.5 \\
$r_k$=16                 & 58.7 & 90.5 & 79.2 & 89.7 \\
$r_v$=16                 & 66.5 & 92.1 & 85.9 & 91.9 \\
$r_o$=16                 & 65.4 & 91.7 & 84.8 & \textbf{92.0} \\
$r_q$=$r_k$=8            & 62.6 & 90.8 & 83.1 & 90.4 \\
$r_k$=$r_v$=8            & 67.7 & 89.5 & 85.8 & 91.7 \\
$r_q$=$r_k$=$r_v$=$r_o$=4& 68.5 & 89.0 & 85.6 & 91.9 \\ \bottomrule
\end{tabular}
\end{table}
Given a limited parameter budget, which types of rank settings yield the best performance?
To answer the above question, we conduct experiments on four tasks of GLUE benchmark, limiting the total number of trainable parameters to 0.79M on RoBERTa-large.
For simplicity and parameter-efficiency considerations, we only apply low-rank adaptation to the attention weights and keep the FFN modules frozen.
This parameter budget is equivalent to $r=8$ if the low-rank adaptation is applied to two types of weight matrices and $r=16$ when applied to one type.
When adapting to the $W^Q$ and $W^V$ weights, we employed the asymmetric initialization method of AILoRA.
For other weight types, the default initialization method of LoRA is used.
The experimental results are summarized in Table~\ref{rank}.
Notably, allocating the entire parameter budget to the $W^K$ weights almost results in the poorest performance, as previously highlighted in LoRA.
Moreover, applying low-rank adaptation to more than one type of weight matrices generally leads to better performance.
At last, applying AILoRA to the $W^Q$ and $W^V$ weights yields the best results, demonstrating the effectiveness of our weight matrices selection.
\section{Conclusion}
In this paper, we investigate the distinct functional roles of the $W^Q$ and $W^V$ projection matrices in the self-attention mechanism and present \textbf{AILoRA}, a novel parameter-efficient fine-tuning method inspired by these insights.
AILoRA introduces a function-aware asymmetric initialization scheme, leveraging the principal components of $W^Q$ and the minor components of $W^V$ to initialize the respective LoRA modules.
This design enables LoRA to better exploit the functional asymmetry of attention parameters, leading to improved downstream performance and faster convergence.
Extensive experiments across diverse model architectures (encoder-only, decoder-only, and encoder-decoder), parameter scales (ranging from 184M to 7B), and downstream tasks (including both NLU and NLG benchmarks) consistently validate the effectiveness and robustness of AILoRA.
\bibliography{anonymous-submission-latex-2026}

\makeatletter
\@ifundefined{isChecklistMainFile}{
  \newif\ifreproStandalone
  \reproStandalonetrue
}{
  \newif\ifreproStandalone
  \reproStandalonefalse
}
\makeatother

\ifreproStandalone
\documentclass[letterpaper]{article}
\usepackage[submission]{aaai2026}
\setlength{\pdfpagewidth}{8.5in}
\setlength{\pdfpageheight}{11in}
\usepackage{times}
\usepackage{helvet}
\usepackage{courier}
\usepackage{xcolor}
\frenchspacing

\begin{document}
\fi
\setlength{\leftmargini}{20pt}
\makeatletter\def\@listi{\leftmargin\leftmargini \topsep .5em \parsep .5em \itemsep .5em}
\def\@listii{\leftmargin\leftmarginii \labelwidth\leftmarginii \advance\labelwidth-\labelsep \topsep .4em \parsep .4em \itemsep .4em}
\def\@listiii{\leftmargin\leftmarginiii \labelwidth\leftmarginiii \advance\labelwidth-\labelsep \topsep .4em \parsep .4em \itemsep .4em}\makeatother

\setcounter{secnumdepth}{0}
\renewcommand\thesubsection{\arabic{subsection}}
\renewcommand\labelenumi{\thesubsection.\arabic{enumi}}

\newcounter{checksubsection}
\newcounter{checkitem}[checksubsection]

\newcommand{\checksubsection}[1]{%
  \refstepcounter{checksubsection}%
  \paragraph{\arabic{checksubsection}. #1}%
  \setcounter{checkitem}{0}%
}

\newcommand{\checkitem}{%
  \refstepcounter{checkitem}%
  \item[\arabic{checksubsection}.\arabic{checkitem}.]%
}
\newcommand{\question}[2]{\normalcolor\checkitem #1 #2 \color{blue}}
\newcommand{\ifyespoints}[1]{\makebox[0pt][l]{\hspace{-15pt}\normalcolor #1}}

\section*{Reproducibility Checklist}

\vspace{1em}
\hrule
\vspace{1em}


\checksubsection{General Paper Structure}
\begin{itemize}

\question{Includes a conceptual outline and/or pseudocode description of AI methods introduced}{(yes/partial/no/NA)}
yes

\question{Clearly delineates statements that are opinions, hypothesis, and speculation from objective facts and results}{(yes/no)}
yes

\question{Provides well-marked pedagogical references for less-familiar readers to gain background necessary to replicate the paper}{(yes/no)}
yes

\end{itemize}
\checksubsection{Theoretical Contributions}
\begin{itemize}

\question{Does this paper make theoretical contributions?}{(yes/no)}
no

	\ifyespoints{\vspace{1.2em}If yes, please address the following points:}
        \begin{itemize}
	
	\question{All assumptions and restrictions are stated clearly and formally}{(yes/partial/no)}
	NA

	\question{All novel claims are stated formally (e.g., in theorem statements)}{(yes/partial/no)}
	NA

	\question{Proofs of all novel claims are included}{(yes/partial/no)}
	NA

	\question{Proof sketches or intuitions are given for complex and/or novel results}{(yes/partial/no)}
	NA

	\question{Appropriate citations to theoretical tools used are given}{(yes/partial/no)}
	NA

	\question{All theoretical claims are demonstrated empirically to hold}{(yes/partial/no/NA)}
	NA

	\question{All experimental code used to eliminate or disprove claims is included}{(yes/no/NA)}
	NA
	
	\end{itemize}
\end{itemize}

\checksubsection{Dataset Usage}
\begin{itemize}

\question{Does this paper rely on one or more datasets?}{(yes/no)}
yes

\ifyespoints{If yes, please address the following points:}
\begin{itemize}

	\question{A motivation is given for why the experiments are conducted on the selected datasets}{(yes/partial/no/NA)}
	yes

	\question{All novel datasets introduced in this paper are included in a data appendix}{(yes/partial/no/NA)}
	yes

	\question{All novel datasets introduced in this paper will be made publicly available upon publication of the paper with a license that allows free usage for research purposes}{(yes/partial/no/NA)}
	yes

	\question{All datasets drawn from the existing literature (potentially including authors' own previously published work) are accompanied by appropriate citations}{(yes/no/NA)}
	yes

	\question{All datasets drawn from the existing literature (potentially including authors' own previously published work) are publicly available}{(yes/partial/no/NA)}
	yes

	\question{All datasets that are not publicly available are described in detail, with explanation why publicly available alternatives are not scientifically satisficing}{(yes/partial/no/NA)}
	yes

\end{itemize}
\end{itemize}

\checksubsection{Computational Experiments}
\begin{itemize}

\question{Does this paper include computational experiments?}{(yes/no)}
yes

\ifyespoints{If yes, please address the following points:}
\begin{itemize}

	\question{This paper states the number and range of values tried per (hyper-) parameter during development of the paper, along with the criterion used for selecting the final parameter setting}{(yes/partial/no/NA)}
	yes

	\question{Any code required for pre-processing data is included in the appendix}{(yes/partial/no)}
	yes

	\question{All source code required for conducting and analyzing the experiments is included in a code appendix}{(yes/partial/no)}
	yes

	\question{All source code required for conducting and analyzing the experiments will be made publicly available upon publication of the paper with a license that allows free usage for research purposes}{(yes/partial/no)}
	yes
        
	\question{All source code implementing new methods have comments detailing the implementation, with references to the paper where each step comes from}{(yes/partial/no)}
	yes

	\question{If an algorithm depends on randomness, then the method used for setting seeds is described in a way sufficient to allow replication of results}{(yes/partial/no/NA)}
	no

	\question{This paper specifies the computing infrastructure used for running experiments (hardware and software), including GPU/CPU models; amount of memory; operating system; names and versions of relevant software libraries and frameworks}{(yes/partial/no)}
	yes

	\question{This paper formally describes evaluation metrics used and explains the motivation for choosing these metrics}{(yes/partial/no)}
	yes

	\question{This paper states the number of algorithm runs used to compute each reported result}{(yes/no)}
	yes

	\question{Analysis of experiments goes beyond single-dimensional summaries of performance (e.g., average; median) to include measures of variation, confidence, or other distributional information}{(yes/no)}
	yes

	\question{The significance of any improvement or decrease in performance is judged using appropriate statistical tests (e.g., Wilcoxon signed-rank)}{(yes/partial/no)}
	yes

	\question{This paper lists all final (hyper-)parameters used for each model/algorithm in the paper’s experiments}{(yes/partial/no/NA)}
	yes

\end{itemize}
\end{itemize}
\ifreproStandalone
\end{document}
\fi

\end{document}